\documentclass[conference]{IEEEtran}
\IEEEoverridecommandlockouts
\usepackage{amsmath,amssymb,amsfonts}
\usepackage{algorithmic}
\usepackage{cite}
\usepackage{graphicx}
\usepackage{textcomp}
\usepackage[bookmarks=false,hidelinks]{hyperref}
\usepackage{eso-pic}
\usepackage{multirow}
\usepackage{footmisc}
\usepackage{enumitem}
\usepackage{tabularx}
\usepackage{threeparttable}

\def\BibTeX{{\rm B\kern-.05em{\sc i\kern-.025em b}\kern-.08em
    T\kern-.1667em\lower.7ex\hbox{E}\kern-.125emX}}
\title{PerKey: A Persian News Corpus for Keyphrase \\Extraction and Generation\\}

\author{

\IEEEauthorblockN{Ehsan Doostmohammadi}
\IEEEauthorblockA{\textit{Graduate Student} \\
\textit{Computational Linguistics Group}\\
\textit{Sharif University of Technology}\\
Tehran, Iran \\
e.doostm72@student.sharif.edu}
\and

\IEEEauthorblockN{Mohammad Hadi Bokaei}
\IEEEauthorblockA{\textit{Assistant Professor} \\
\textit{Information Technology Department}\\
\textit{ICT Research Institute}\\
Tehran, Iran \\
mh.bokaei@itrc.ac.ir}
\and

\IEEEauthorblockN{Hossein Sameti}
\IEEEauthorblockA{\textit{Associate Professor} \\
\textit{Computer Engineering Department}\\
\textit{Sharif University of Technology}\\
Tehran, Iran \\
sameti@sharif.edu}
}

\IEEEoverridecommandlockouts
\IEEEpubid{\makebox[\columnwidth]{978-1-5386-8274-6/18/\$31.00 ©2018 IEEE \cite{8661095} \hfill} \hspace{\columnsep}\makebox[\columnwidth]{ }}

\begin{document}
 \AddToShipoutPictureFG*{
                \AtPageUpperLeft{\put(0,-10){\makebox[\paperwidth][l]{2018 9th International                 Symposium on Telecommunications (IST'2018)}}}
                                                           }
\maketitle

\begin{abstract}
Keyphrases provide an extremely dense summary of a text. Such information can be used in many Natural Language Processing tasks, such as information retrieval and text summarization. Since previous studies on Persian keyword or keyphrase extraction have not published their data, the field suffers from the lack of a human extracted keyphrase dataset.
In this paper, we introduce PerKey\footnote{\href{https://github.com/edoost/perke}{https://github.com/edoost/perkey}\label{refnote}}, a corpus of 553k news articles from six Persian news websites and agencies with relatively high quality author extracted keyphrases, which is then filtered and cleaned to achieve higher quality keyphrases. The resulted data was put into human assessment to ensure the quality of the keyphrases. We also measured the performance of different supervised and unsupervised techniques, e.g. TFIDF, MultipartiteRank, KEA, etc. on the dataset using precision, recall, and \texorpdfstring{F\textsubscript{1}}\normaltext -score.
\end{abstract}


\section{Introduction}
Keyphrases are single words or sequences of words that express the main topics discussed in a piece of text. High quality keyphrases can help the categorization and retrieval of data, which can be used in many Natural Language Processing (NLP) tasks, including information retrieval, text summarization, text categorization, question answering, and opinion mining.

Keyphrase extraction is quite a well-known task in NLP, the purpose of which is to extract keyphrases from a given text. Although a considerable progress has been made, the effectiveness of the task of keyphrases extraction is still far from many other problems in NLP. Methods used to extract keyphrases vary from unsupervised approaches, e.g. statistical and graph-based scoring methods, to supervised approaches, e.g. classification/sequence labelling methods.
Since supervised approaches require labelled data, most of the approaches to the problem of keyphrases extraction are unsupervised. Unsupervised techniques tend to perform better in general and supervised approaches in specific fields. Additionally, state-of-the-art methods, i.e. artificial neural networks, require training on labelled data. The problem with both unsupervised and supervised approaches is that they are unable to extract implicit keyphrases. Not all the keyphrases are always explicitly mentioned in the text, but some of them are implicitly spoken of. To extract such keyphrases we need generative models, e.g. sequence to sequence methods as in neural machine translation.

For the abovementioned reasons, a need for a high quality keyphrase corpus is strongly felt in Persian natural language processing. As hand-annotating data is not easily feasible for large corpora, we decided to gather author-keyphrase-annotated data from news websites and agencies with relatively high quality keyphrases. This method is precedented in the task of keyphrase extraction from tweets \cite{zhang2016keyphrase} and keyphrase generation from scientific papers \cite{meng2017deep}, the latter collecting 567,830 papers in the domain of computer science from multiple websites. As of our work, we assessed more than 30 news websites and agencies and chose 6 of them having the highest quality keyphrases. After collecting the data, we did some cleaning and preprocessing to improve the quality of data which resulted in 553,111 news articles, their keyphrases, summaries, and some other information. The resulted data was put into human assessment, in a way which will be described, to ensure the quality of the keyphrases. The dataset is available at this project's GitHub repository \footref{refnote}.

In the next section we discuss the different approaches to keyphrase extraction and generation in English and Persian. In section III, the dataset of news articles and their keyphrases will be introduced in details and the method of collecting and cleaning them will be discussed. In section IV different methods tested on the dataset are described in details. In section V and VI, the results are represented and discussed. We conclude this work and discuss the future work in section VII.

\section{Previous Work}
\subsection{Work on English}
Although a substantial number of approaches to keyphrase extraction are unsupervised\cite{hasan2014automatic,brin1998anatomy,liu2009clustering,liu2010automatic}, recent supervised algorithms outperform the conventional unsupervised ones\cite{zhang2016keyphrase,meng2017deep}. Achieving such performance requires a large set of training data which is not achievable without a great deal of human resources, time, and financial cost. Instead, one can make use of already tagged data in the internet, even though they are not as accurate as such gold dataset. Such approach is taken in keyprhase extraction from tweets \cite{zhang2016keyphrase} and keyphrase generation in the field of scientific papers \cite{meng2017deep}. Some other publicly available datasets in English are \cite{medelyan2009human,kim2010semeval} for scientific papers, \cite{wan2008single} for news articels, and \cite{yih2006finding} for web pages, none of which containing more than 1,000 samples, which makes them unsuitable for training neural networks.

\subsection{Work on Persian}
Literature of keyphrase/keyword extraction from Persian texts is quite minimal. Most of the papers on Persian take an unsupervised approach and non of them has released their data.

Mohammadi and Analoui \cite{ACCSI13_067}, after removing stop words and stemming, extracts keywords using TF, TTF (Total Term Frequency), and DF matrices and fuzzification.

Khozani and Bayat \cite{khozani2011specialization}, after stemming and removing stop words, uses $W(t_k) = \frac{tf(d_i, t_k) \times N}{df(t_k)}$ to obtain TFIDF weights of the words in documents. $W$ being the weight,  $t_k$ the $k$'th token, $tf$ the term frequency, $d_i$ the $i$'th document, N number of tokens in the current document, and $d_f$ the document frequency. After calculating the weights of the tokens, the weight of the bi-token keyphrases is calculated using a co-occurrence matrix.

Kian and Zahedi \cite{kian2013improving} uses attention attractive strings
to improve keyword extraction from 800 documents from Hamshahri
news collection \cite{aleahmad2009hamshahri}. Three to seven keyphrases were assigned to each document, the length of each keyphrase ranging from two to four words. The best result, an \texorpdfstring{F\textsubscript{1}}\normaltext of $40.23$, is achieved using attention attractive strings and training on 400 documents.

\section{Dataset Description}

We assessed more than 30 news websites and agencies and chose 6 of them having the highest quality keyphrases. The dataset comprises 553,111 news articles crawled from 6 news websites and agencies which provided high-quality keyphrases. Number of news articles crawled from each website/agency can be found in table I.

\begin{table}[ht]
\caption{Number of Articles Based on Website/Agency}
\label{Table:num_news}
\begin{center}
\begin{tabular}{|c|c|c|}
\hline
website/agency & \# of articles & \% of total \\
\hline
\hline
Khabaronline & 165,839 & 29.98\% \\
\hline
Fararu & 161,588 & 29.21\% \\
\hline
Ilna & 145,420 & 26.29\% \\
\hline
Mashreq & 33,621 & 6.07\% \\
\hline
Alef & 33,621 & 6.07\% \\
\hline
Aftab & 1,517 & 0.27\% \\
\hline
total & 553,111 & 100\% \\
\hline
\end{tabular}
\end{center}
\end{table}

The dataset is stored in JSON format, each news article containing the following information:

\noindent
\begin{verbatim}
{title, body, summary, keyphrases, category, url}
\end{verbatim}

After character and word normalization, the articles with bodies containing less than 40 and more than 500 tokens were removed. This was to make sure that the articles are mostly news, as articles longer than 500 words tend to be interview or news analysis. More information on number of tokens in the body of each article can be found in table II. 

\begin{table}[ht]
\caption{Number of Articles Based on Number of Tokens in them}
\label{Table:len_body}
\begin{center}
\begin{tabular}{|c|c|c|}
\hline
\# of tokens & \# of articles & \% of total \\
\hline
\hline
40-100 & 123,913 & 22.40\% \\
\hline
100-150 & 96,919 & 17.52\% \\
\hline
150-200 & 81,857 & 14.79\% \\
\hline
200-250 & 66,738 & 12.06\% \\
\hline
250-300 & 54,119 & 9.78\% \\
\hline
300-350 & 42,826 & 7.74\% \\
\hline
350-400 & 34,457 & 6.23\% \\
\hline
400-450 & 28,194 & 5.09\% \\
\hline
450-500 & 24,088 & 4.35\% \\
\hline
total & 553,111 & 100\% \\
\hline
\end{tabular}
\end{center}
\end{table}

We also removed news articles with less than 2 keyphrases. It appears that most of the articles (65.11\%) contain 2 or 3 keyphrases. More information on this can be found in table III. 

\begin{table}[ht]
\caption{Number of Articles Based on Number of Keyphrases}
\label{Table:len_body}
\begin{center}
\begin{tabular}{|c|c|c|}
\hline
\# of keyphrases & \# of articles & \% of total \\
\hline
\hline
2 & 157,466 & 28.46\% \\
\hline
3 & 202,748 & 36.65\% \\
\hline
4 & 81,500 & 14.73\% \\
\hline
5 & 56,278 & 10.17\% \\
\hline
6 & 7,060 & 1.27\% \\
\hline
7 & 2,883 & 0.52\% \\
\hline
8 & 1,382 & 0.25\% \\
\hline
9 & 1,032 & 0.18\% \\
\hline
9+ & 42,762 & 7.73\% \\
\hline
total & 553,111 & 100\% \\
\hline
\end{tabular}
\end{center}
\end{table}

Table IV shows number of keyphrases based on the number of characters in them. It appears that most of the keyphrases contain something between 5 to 15 characters. News articles containing keyphrases with less than 3 characters length were removed.

\begin{table}[ht]
\caption{Number of Keyphrases Based on the Number of Characters in them}
\label{Table:len_body}
\begin{center}
\begin{tabular}{|c|c|c|}
\hline
\# of characters & \# of keyphrases & \% of total \\
\hline
\hline
3-5 & 276,141 & 13.52\% \\
\hline
5-10 & 830,036 & 40.65\% \\
\hline
10-15 & 574,945 & 28.16\% \\
\hline
15-20 & 233,032 & 11.41\% \\
\hline
20-25 & 82,354 & 4.03\% \\
\hline
25-30 & 25,771 & 1.26\% \\
\hline
30-35 & 14,955 & 0.73\% \\
\hline
35-40 & 4,392 & 0.21\% \\
\hline
total & 2,041,626 & 100\% \\
\hline
\end{tabular}
\end{center}
\end{table}

Table V shows that most of the keyphrases contain either 1 or 2 tokens. News articles containing keyphrases with more than 7 tokens in them were removed.

\begin{table}[ht]
\caption{Number of Keyphrases Based on the Number of Tokens in them}
\label{Table:len_body}
\begin{center}
\begin{tabular}{|c|c|c|}
\hline
\# of tokens & \# of keyphrases & \% of total \\
\hline
\hline
1 & 943,485 & 46.21\% \\
\hline
2 & 746,148 & 36.54\% \\
\hline
3 & 222,339 & 10.90\% \\
\hline
4 & 89,746 & 4.40\% \\
\hline
5 & 25,933 & 1.27\% \\
\hline
6 & 10,800 & 0.52\% \\
\hline
7 & 3,175 & 0.15\% \\
\hline
total & 2,041,626 & 100\% \\
\hline
\end{tabular}
\end{center}
\end{table}

Finally, it appears that around 717k of the keyphrases, which constitute 35.13\% of all keyphrases, are not present in the body of the news articles. This number shows that the dataset is suitable for the task of keyphrase generation, as well as extraction. More on this in table VI.

\begin{table}[ht]
\caption{Number of Absent and Present Keyphrases}
\label{Table:absent_present}
\begin{center}
\begin{tabular}{c||c|c}
& \# of keyphrases & \% of total \\
\hline
\hline
present & 1,324,305 & 64.86\% \\
\hline
absent & 717,321 & 35.13\% \\
\hline
total & 2,041,626 & 100\% \\
\end{tabular}
\end{center}
\end{table}

The dataset is divided to three parts, 25,000 for test set, 25,000 for validation set, and 503,111 for training set. The dataset can be downloaded from this project's GitHub repository referenced in the first page. To ensure the quality of the keyphrases, we conducted a survey, asking 5 participants to rate the precision and recall of the reference keyphrases. Participants were asked to answer two questions from 1 to 5 for 50 randomly chosen news articles, the two questions being:
\begin{itemize}
\item How much are the keyphrases related to the subjects discussed in the news article?
\item To what extent are the subjects discussed in the article reflected in the keyphrases?
\end{itemize}

The average of the answers to the first question was $4.592$ and to the second $3.980$, the harmonic mean of which being $4.264$, which ensures the quality of the keyphrases.

\section{Methodology of the Tested Algorithms}
In this section the approaches to extract keyphrases from news articles will be discussed. We can divide the approaches into two main models, unsupervised and supervised. All the models are implemented using pke (an open source python-based keyphrase extraction toolkit) \cite{boudin:2016:COLINGDEMO}.

\subsection{Unsupervised Models}
The most important merit of unsupervised models is that they can work in a new area or language with little or no adjustments. We can divide the unsupervised techniques into two subsets: statistical models and graph-based models.

\subsubsection{Statistical Models}
Statistical models are models that select candidates based on their statistical features. 

\begin{itemize}
\item TFIDF \cite{kim2013automatic} calculates the importance of $n$-gram phrases in a corpus of documents. Here, IDF is calculated using below formula:

\[
idf = \log (1 + \frac{N}{n_t})
\]

$N$ being the total number of documents in the corpus, and $n_t$ the number of documents containing the term $t$.

\item KPMiner \cite{el2009kp} works in three phases: (1) candidate selection, (2) candidate weight calculation, and (3) keyphrase refinement. The first phase, candidate selection, follows three heuristics: (a) the phrases should not be separated by punctuation or stop word, (b) the resulted phrases should have appeared at least $n$ time(s) in the document (LASF: least allowable seen frequency), and (c) as phrases happening after a certain threshold in a document are very rarely keyphrases, they will be cut off. The candidate weight calculation phase is done using $w_{ij} = tf_{ij} \times idf \times B_i \times P_f$ where $w_{ij}$ is weight of term $t_j$ in document $D_i$, $tf_{ij}$ is the frequency of term $t_j$ in document $D_i$, $idf$ is $log_2 (N/n)$ where $N$ is the number of documents in the collection and $n$ is number of documents where term $t_j$ occurs at least once, $P_f$ is the position factor, and $B_i$ is the boosting factor associated with document $D_i$ calculated using below formula:

\[
B_d =  \frac{|Nd|}{(|P_d| \times \alpha)}
\]

where if $B_d > \sigma$ then $B_d = \sigma$ and $|P_d|$ is the number of candidate terms whose lengths exceed one in the document and $\alpha$ and $\sigma$ are hyperparameters. Finally, the weights of the top $n$ keyphrases are refined by seeing if any of them is a sub-phrase of another. If so, its count is decremented
by the frequency of the term of which it is a part and weights are recalculated. Here the cutoff threshold and LASF were tuned on the development set and were set to $250$ words and $3$, respectively. We left $\alpha$ and $\sigma$ to their default values, i.e. $2.3$ and $3.0$ respectively.

\item YAKE \cite{campos2018text,campos2018yake} is a multilingual online single-document keyphrase extraction tool which relies only on statistical features, and a list of stopwords, to extract keyphrases from unlabelled data. The process comprises six steps: (1) preprocessing: splitting the tokens based on word boundaries; (2) feature extraction, consisting of:
\begin{enumerate}
\item Casing: regards the casing aspect of a word;
\item Word Positional: values words occurring at the beginning;
\item Word Frequency: scores more the words occurring more often;
\item Word Relatedness to Context: computes the number of different terms that occur in the context of the candidate word on the basis that the number of different terms that co-occur with the candidate word has a direct relationship with the meaninglessness of it. In other words, the more diverse the context of the candidate, the more meaningless it is;
\item Word DifSentence: counts the number of times a candidate word appears within different sentences.

(3) Individual terms are scored using the abovementioned features; (4) Candidate keywords list are generated to the size of the window (a hyperparameter) and the scores are calculated as below such that the smaller the score the better the keyword will be:

\[
S(kw) = \frac{\prod\limits_{w \in kw} S(w)}{TF(kw) \times (1 + \sum\limits_{w \in kw} S(w))}
\]

where S(kw) is the score of the candidate phrase calculated as the product of the scores of the constituent words normalized by the length and TF of the keyphrase, such that longer n-grams will not be scored higher and less frequent ones will not be penalised. (5) Data is deduplicated using Levenshtein distance, and finally (6) the keyphrases are ranked such that the lower the $S(kw)$ score, the more important the
keyword will be.

\end{enumerate}
\end{itemize}

\subsubsection{Graph-based Models}
These approaches typically consist of two steps: (1) building a
graph representation of the document with words as nodes and semantic relation between them as edges; (2) ranking nodes
using a graph-theoretic measure and using the
top-ranked ones as keyphrases.

\begin{itemize}
\item SingleRank \cite{wan2008collabrank} extracts candidates from sequences of nouns and adjectives and builds a word-based graph of the document and scores them using TextRank \cite{mihalcea2004textrank}.

\item TopicRank \cite{bougouin2013topicrank} improves SingleRank by grouping lexically similar candidates into topics and ranking topics. After extracting sequences of nouns and adjectives as noun phrases, TopicRank clusters them into topics using a hierarchical agglomerative clustering algorithm with average linkage. Next, similar phrases are removed and the graph is built and weighted and final keyphrases are selected from first occurring candidates of the $n$ highest ranked topics.

\item Multipartite Rank \cite{boudin2018unsupervised} encodes topical information within a multipartite graph structure. This method works in two steps like other graph-based approaches. However, it also has a middle step in which weights are adjusted in order to capture position information. In the first step, keyphrase candidates are selected from what we define as noun phrase. As we are dealing with Persian language in this paper, the pattern for noun phrase would be the sequence of nouns with zero or more adjectives (/Noun+Adj*/). Then, based on the stemmed form of the words, the topics are found using hierarchical agglomerative
clustering with average linkage. The candidate similarity is calculated as below:

\[
sim(c_i, c_j) = \frac{|stems(c_i) \cap stems(c_j)|}{|stems(c_i) \cup stems(c_j)|}
\]

A directed multipartite graph is built, the nodes of which are keyphrase candidates that are connected only if they belong to different topics and edges are weighted according to the distance between two candidates:

\[
w_{ij} = \sum_{c_i \in t_i} \sum_{c_j \in t_j} dist(c_i, c_j)
\]

$c$ being the candidate and $t$ the topic. Next, the most representative keyphrase candidates
for each topic are selected according to their position in the document. In other words, candidates occurring at the beginning are promoted
to the other candidates of the same topic. Finally, keyphrase candidates are
ordered by TextRank ranking algorithm \cite{mihalcea2004textrank}:

\[
S(c_i) = (1 - \lambda) + \lambda \; . \sum_{c_j \in \mathcal{I}(c_i)} \frac{w_{ij} . S(c_j)} {\sum\limits_{c_k \in \mathcal{O}(c_j)} w_{jk}}
\]

where $\mathcal{I}(c_i)$ is the set of predecessors of $c_i$, $\mathcal{O}(c_j)$
is the set of successors of $c_j$ , and $\lambda$ is a damping
factor set to $0.85$.
\end{itemize}
\subsection{Supervised Model}
Supervised models need annotated data and it takes time to train them. However, they often show superior performance and effectiveness comparing to unsupervised techniques.
\begin{itemize}
\item KEA \cite{witten2005kea} calculates two features for each sample: TFIDF and first occurrence. TFIDF is calculated using:

\[
tfidf = \frac{freq(P,D)}{size(D)} \times -\log_2 \frac{df(P)}{N}
\]

where $P$ stands for phrase and $size()$ is the number of words. The second feature, first occurrence, is calculated as the number of words that precede the
phrase’s first appearance, divided by the number of words in the document. It results in a number between $0$ and $1$ indicating the percentage of the document that precedes phrase's first appearance. After discretization of the values of these two features for gold keyphrases, the probability of a phrase being a keyphrase is calculated as:

\[
P(yes)= \frac{Y}{Y + N} \times P_{tfidf} (t | yes) \times P_{dist} (d | yes)
\]

where $t$ is the feature value for $tfidf$ and $d$ for $distance$. Finally, the probability of keyphraseness is calculated as:

\[
P(keyphraseness) = \frac{P(yes)}{P(yes) + P(no)}
\]

\end{itemize}

\section{Results}
We conducted an empirical study on 7 models, the results of which are shown in table VII, @5 meaning the results on the top five keyphrases and @10, top ten. We also repeated the experiment this time only on news articles with 3 and more keyphrases. The results are available in table VIII. The best results are in bold font, second best are underlined, and the third best results are in italics.

\begin{table}[ht]
\caption{Empirical Results on All the Data}
\label{Table:absent_present}
\begin{center}
\begin{tabular}{|c|c|c|c|c|c|c|}
\hline
Method & P@5 & R@5 & F\textsubscript{1}@5 & P@10 & R@10 & F\textsubscript{1}@10 \\
\hline
\hline
TFIDF & \emph{.1459} & \underline{.1959} & \emph{.1673} & \emph{.1036} & \underline{.2758} & \emph{.1507} \\
\hline
KPM. & \bf{.1625} & \emph{.1843} & \underline{.1727} & \bf{.1404} & \emph{.2371} & \bf{.1763} \\
\hline
YAKE & .0622 & .0806 & .0702 & .0568 & .1459 & .0818 \\
\hline
S.Rank & .0500 & .0767 & .0605 & .0552 & .1605 & .0822 \\
\hline
T.Rank & .0846 & .1189 & .0989 & .0566 & .1541 & .0828 \\
\hline
M.Rank & .0922 & .1268 & .1068 & .0652 & .1772 & .0953 \\
\hline
KEA & \underline{.1564} & \bf{.2125} & \bf{.1802} & \underline{.1110} & \bf{.2990} & \underline{.1620} \\
\hline
\end{tabular}
\end{center}
\end{table}

The results in table VII and also table VIII show that supervised method, KEA, performs the best, and after that, statistical methods tend to perform better in comparison to graph-based ones.

\begin{table}[ht]
\caption{Empirical Results on News Articles with at Least Three Keyphrases}
\label{Table:absent_present}
\begin{center}
\begin{tabular}{|c|c|c|c|c|c|c|}
\hline
Method & P@5 & R@5 & F\textsubscript{1}@5 & P@10 & R@10 & F\textsubscript{1}@10\\
\hline
\hline
TFIDF & \emph{.1724} & \underline{.2060} & \emph{.1877} & \emph{.1216} & \underline{.2877} & \emph{.1710} \\
\hline
KPM. & \bf{.1900} & \emph{.1948} & \underline{.1924} & \bf{.1632} & \emph{.2513} & \bf{.1979}\\
\hline
YAKE & .0726 & .0820 & .0770 & .0658 & .1481 & .0911 \\
\hline
S.Rank & .0532 & .0671 & .0594 & .0623 & .1533 & .0886 \\
\hline
T.Rank & .0986 & .1208 & .1086 & .0665 & .1583 & .0937 \\
\hline
M.Rank & .1093 & .1319 & .1196 & .0771 & .1835 & .1086 \\
\hline
KEA & \emph{.1837} & \bf{.2226} & \bf{.2013} & \underline{.1300} & \bf{.3115} & \underline{.1835} \\
\hline
\end{tabular}
\end{center}
\end{table}

Comparing the results in table VII and VIII shows that keyphrases in news articles with fewer keyphrases do not cover all the subjects discussed in the news article, hence are of low quality.

\section{Discussion}
Experimental results show that the supervised method, as opposed to what expected, cannot outperform KPMiner. Apparently, this method is not powerful enough to benefit from such training data which emphasizes the need for more powerful state-of-the-art techniques. The tables also show that statistical unsupervised approaches outperform graph-based ones. That is also the case in English standard datasets \cite{kim2010semeval}.

Another experiment we performed, was removing news articles having fewer than three keyphrases from the data and repeating the evaluations. Supervised and unsupervised techniques showed higher precision, recall, and \texorpdfstring{F\textsubscript{1}}\normaltext -score when we eliminated those news articles. One reason could be the higher quality of the news articles with more keyphrases, in general, also supported by the quality assessment performed by human on the dataset. 

\section{Conclusion and Future Work}
In this paper, we tested 7 keyphrase extraction models on two kinds of data. Limiting the news articles to those that have more keyphrases not only keeps news articles with more representative keyphrases, but also results in better performance in training of the supervised technique, i.e. KEA. Such results are promising for training state of the art supervised approaches, e.g. sequence labelling with artificial neural networks, as they performance drop when learning from very sparse data. We also observed that 35.13\% of all the keyphrases are absent from news body. Absent keyphrases are prevalent in real texts as the subjects discussed are not always mentioned explicitly in them. Tackling such problem requires generative models, e.g. sequence to sequence models.

It is also important to improve evaluation schemes. Current evaluation convention in keyphrase extraction, i.e. precision, recall, and \texorpdfstring{F\textsubscript{1}}\normaltext -score, is too much strict, because different words can mean the same thing and refer to the same subject. \cite{hasan2014automatic} suggests some improvements to the criteria, for instance, using machine translation evaluation techniques, e.g. BLEU and Rouge, to achieve a more realistic evaluation for the task of keyphrase extraction and generation.

\bigskip

\bibliographystyle{IEEEtran}
\bibliography{bibliography}

\end{document}